\documentclass[letterpaper, 10 pt, conference]{ieeeconf}  

\usepackage{graphicx}

\usepackage[caption=false,font=normalsize,labelfont=sf,textfont=sf]{subfig}
\usepackage{cite}
\usepackage{amssymb}

\IEEEoverridecommandlockouts                              

\title{\LARGE \bf
Adaptive Motion Planning for Multi-fingered Functional Grasp via Force Feedback
}

\author{Dongying Tian$^{1,2}$, Xiangbo Lin$^{1}$ and Yi Sun$^{1}$
	\thanks{This work was supported by the National Natural Science Foundation of China [grant numbers 61873046, U1708263].\emph{(Corresponding author:Yi Sun)}}
\thanks{$^{1}$Dongying Tian, Xiangbo Lin and Yi Sun are with
	the School of Information and Communication and Engineering,
	Dalian University of Technology, Dalian, 116024, China. (email:
	tiandongying@sia.cn, linxbo@dlut.edu.cn,  lslwf@dlut.edu.cn.)}
\thanks{$^{2}$Dongying Tian is also with Shenyang Institute of Automation, Chinese Academy
	of Sciences, Shenyang 110016, China.}
}

\begin{document}

\maketitle
\thispagestyle{empty}
\pagestyle{empty}

\begin{abstract}

	Enabling multi-fingered robots to grasp and manipulate objects with human-like dexterity is especially challenging during the dynamic, continuous hand-object interactions. Closed-loop feedback control is essential for dexterous hands to dynamically finetune hand poses when performing precise functional grasps. This work proposes an adaptive motion planning method based on deep reinforcement learning to adjust grasping poses according to real-time feedback from joint torques from pre-grasp to goal grasp. We find the multi-joint torques of the dexterous hand can sense object positions through contacts and collisions, enabling real-time adjustment of grasps to generate varying grasping trajectories for objects in different positions. In our experiments, the performance gap with and without force feedback reveals the important role of force feedback in adaptive manipulation. Our approach, utilizing force feedback, preliminarily exhibits human-like flexibility, adaptability, and precision.
\end{abstract}

\section{INTRODUCTION}

Enabling multi-fingered robots to grasp and manipulate objects with human-like dexterity is a challenging task that typically involves two steps: grasp synthesis\cite{newbury2023deep} and motion planning. Previous studies have primarily focused on synthesizing static grasps for specific objects \cite{grady2021contactopt,zhu2021toward}. However, specifying only the static grasp configuration is insufficient, as grasp and manipulation involve continuous hand-object interaction. Errors in the hand or object pose can result in collisions, complicating motion planning and trajectory optimization. Multi-fingered robot hands face even greater difficulty as the possible interaction modes grow exponentially with the number of hand links and object contact points. This is particularly true for functional grasps, such as those required for dexterous tool use, which require accurate touching of specific functional parts, like  the nozzle of a watering can or the button of an electric drill. Closed-loop feedback control based on sensory observations is therefore essential to dynamically updating and adapting the grasping process to disturbances and errors.

Recently, deep reinforcement learning has emerged as an approach to controlling complex dynamical systems, especially in high-dimensional dexterous manipulation \cite{she2022learning,christen2022d}. These approaches often rely on human demonstrations \cite{rajeswaran2017learning} or assume precise knowledge of the object's pose throughout the grasp and manipulation process  \cite{christen2022d}. However, partial occlusion of the object due to its own shape or the manipulator can cause errors in object shape and pose estimation, leading to failed grasps. In such situations, force feedback with dynamic responses to disturbances is essential to compensate for the object's shape and pose mismatch. Previous studies \cite{merzic2019leveraging,wu2019mat,koenig2022role} have shown that learning grasping policies under tactile and force feedback can adapt to environmental uncertainty, thereby improving the grasp success rate for grippers. However, this ability is still challenging to transfer to high-DOF multi-fingered robot hands. Until recently, the work of \cite{liang2021multifingered} trained a five-fingered hand to learn robust motions and stable grasps using force feedback without visual sensing in a pick-up task. 

Our work aims to go beyond the simple pick-and-place task and achieve functional grasp for multi-fingered hands. Functional grasp involves task-oriented grasping on specific functional regions of objects, enabling the completion of manipulations that are both stable and functional. For instance, the index finger presses the shutter of a camera to take photos. In this work, we formulate functional grasp as a dynamic motion planning process, from pre-grasp to goal grasp, highlighting the challenges in contact-rich functional grasp, which requires high precision and robustness against uncertainty. During this process, the object often fails to remain steady due to errors in object and hand pose, resulting in unsynchronized finger contact and the combined translation and rotation of objects in 3D space. Given the vision limitations due to occlusion by the object or manipulator, we study the effectiveness of force feedback to handle uncertainty and design a force-aware path planning method to adapt to the manipulation of the object. This is a more challenging task as there are neither full demonstration trajectories nor visual sensing during grasping.

The contributions of this paper can be summarized as follows:
\begin{itemize}
	\item We propose an adaptive motion planning method for multi-fingered functional grasp via joint torque feedback, which is constantly updated and dynamically adapted to object uncertainty in the absence of visual feedback.
	\item We build a reinforcement learning model in which joint torque feedback is considered as part of the state in the Markov Decision Process, which enables the robot to learn the skill of grasp through trial and error under pose uncertainty.
	\item We conduct functional grasp experiments in simulation and observe a significant effect of force feedback in generating smooth trajectories for functional grasps.
\end{itemize}

\section{Related Work}

\subsection{Analytical Motion Planning}

Nowadays, there are efficient search and optimization algorithms developed to solve motion planning problems \cite{lozano2014constraint}. However, dexterous hand motion planning is still challenging due to the high-dimensional freedom of motion and the complexity of making and breaking contacts between hands and objects \cite{orthey2021sparse}. In 2006, Yunt et al. \cite{yunt2006trajectory} proposed a Contact-Implicit Trajectory Optimization (CITO) method, enabling the numerical solution of infinite-dimensional optimal control problems. To achieve effective solutions and reasonable computation times in dexterous manipulation tasks, the CITO method \cite{posa2014direct} is further enhanced with a discrete contact sequence planner \cite{chen2021trajectotree}. Orthey et al. improve the planning performance on high-dimensional planning problems by using multilevel abstractions to simplify state spaces \cite{orthey2020multilevel} and generalize sparse roadmaps to multilevel abstractions afterwards \cite{orthey2021sparse}. The CMGMP algorithm \cite{cheng2022contact} utilizes automatically enumerated contact modes of environment-object contacts to guide the tree expansions during the search, generating hybrid motion plans including both continuous state transitions and discrete contact mode switches. Another proposed method \cite{pang2023global} enables efficient global motion planning for highly contact-rich and high-dimensional systems. However, analysis methods are usually based on conditional assumptions, and the models are complex and difficult to reproduce.

\subsection{Learning Based Motion Planning}
\begin{figure*}[t]
	\centering
	\includegraphics[width=5.4in]{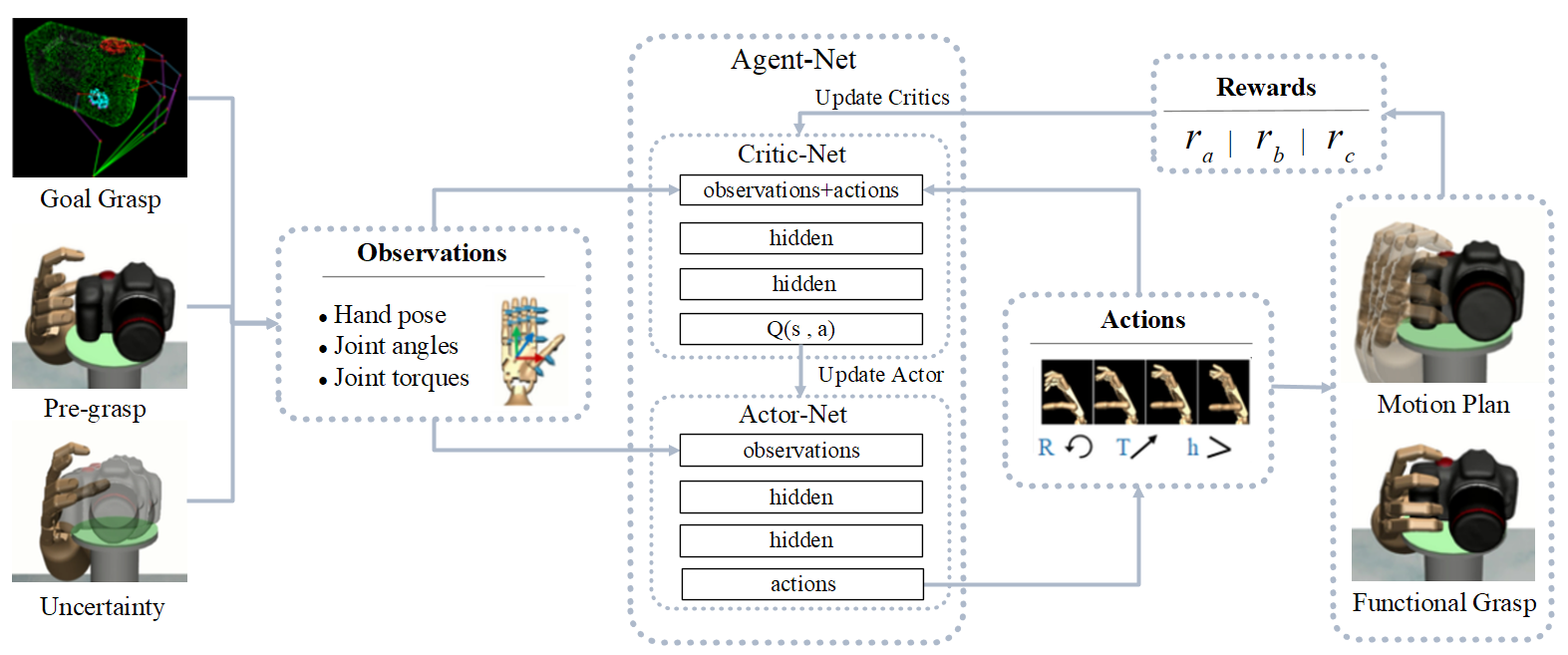}
	\caption{In the scenario where there is uncertainty in the initial position of an object and the pose of the object is unknown during the grasping process, an agent uses hand perception to obtain information about the hand-object interaction state, and makes action decisions accordingly. The rewards obtained from the dexterous hand executing actions will guide the update of the critic-net and further update the actor-net.}
	\label{fig_1}
\end{figure*}
Currently, there have been a few studies that utilize techniques such as CVAE \cite{ye2023learning}, trajectory imitation learning \cite{chen2022dextransfer}, or auto-regressive network architecture \cite{taheri2022goal} to generate grasping paths. However, the majority of research in this area relies on deep reinforcement learning, which has demonstrated outstanding performance in sequential decision-making problems \cite{patel2022learning,dasari2023learning,jain2019learning,mandikal2022dexvip,xu2023unidexgrasp,christen2022d}. To facilitate exploration and reduce sampling complexity, it is common to incorporate prior knowledge such as expert demonstrations \cite{patel2022learning,jain2019learning}, contact maps \cite{mandikal2022dexvip}, or grasping poses \cite{dasari2023learning,xu2023unidexgrasp}. For example, learning efficiency can be enhanced by using a small amount of teaching data \cite{jain2019learning}, or by directly acquiring operational experience from human hand manipulation videos \cite{patel2022learning}. Moreover, a binary affordance map has been employed \cite{mandikal2022dexvip} to guide the agent towards functional grasp regions on the object. Sudeep et al. \cite{dasari2023learning} find that the pregrasp finger pose is crucial for successful behavior learning. Our research, similar to \cite{xu2023unidexgrasp,christen2022d}, utilizes goal grasp poses. However, unlike previous approaches, we specifically focus on functional grasping. In order to accomplish subsequent operational tasks, the hand configuration is typically more complex and requires precise execution, rather than solely picking-up the object.

For the hand-object interaction phase we are studying, not only is there initial position uncertainty in the object, but the object can also be perturbed during the operation. Some studies assume that the object's pose is completely known throughout the entire grasping process, which does not align with real-world scenarios \cite{merzic2019leveraging,liu2023dexrepnet,christen2022d,vulin2021improved,dasari2023learning,patel2022learning}. \cite{xu2023unidexgrasp} addresses this issue by employing a teacher-student distillation approach, where the reinforcement learning of operational strategies is initially performed under the assumption of known object states and later imitated when the object state is unknown. In practice, vision is advantageous in guiding dexterous hands to approach and grasp objects. However, once the hand and object come into contact, precise object position and the interaction forces between the hand and object are difficult to obtain through vision due to occlusion \cite{hu2022physical}. As a result, purely vision-guided grasping often leads to imprecise and jarring operations \cite{liu2023dexrepnet,qin2023dexpoint}.

\subsection{Force Feedback-guided Motion Planning}

Based on human operational experience, real-time force feedback is indispensable for guiding smooth and dexterous manipulation actions. Current research has focused on exploring force feedback on two-finger \cite{vulin2021improved} and three-finger grippers \cite{merzic2019leveraging,wu2019mat,koenig2022role}. These studies emphasize the potential of incorporating tactile feedback in robotic grasping, but it remains highly challenging to extend these techniques to five-finger dexterous hands. Currently, most research in this area has focused on fingertip tactile feedback \cite{sundaralingam2019robust,kumar2019contextual,matak2022planning,liang2021multifingered}. In \cite{sundaralingam2019robust}, fingertip tactile information is mapped to force signals for force feedback control. \cite{kumar2019contextual} utilizes fingertip tactile feedback to compensate for the reduction in geometric information caused by coarse bounding boxes and uncertainties in pose estimation. \cite{matak2022planning} uses fingertip tactile feedback to adapt to online estimates of the object's surface, correcting errors in the initial plan. Liang et al. \cite{liang2021multifingered} propose a fusion of binary contact information from the fingertip, torque sensors, and robot proprioception (joint positions), which shows promise for achieving more robust and stable grasping. Additionally, \cite{jain2019learning} demonstrates that touch sensors capable of sensing on-off contact events enable faster learning in tasks with occlusion.

However, there exists a significant disparity between the perception capabilities of tactile sensors and human touch \cite{li2020review}. When tactile sensors are placed on the finger surface, due to the limitations of the mechanical structure, they usually provide limited low-precision tactile perception. Additionally, frequent hand-object interactions can lead to damage to tactile sensing devices. In contrast, the robot's body typically offers joint torque sensing information, which dynamically changes in response to dexterous hand motion, contact, and collision, providing valuable insights for manipulation. Though several studies have utilized joint torque information \cite{xu2023unidexgrasp,chen2022towards,liang2021multifingered}, it remains uncertain whether dexterous hands have obtained the control skills to utilize joint torque feedback for performing operations. In our research, we solely rely on joint torque feedback to guide the dexterous hand, thereby revealing the potential value of joint torque feedback.

\section{PROBLEM FORMULATION OF FUNCTIONAL GRASP PLANNING}

This work aims to plan a motion trajectory for a functional grasp that is not only stable but also functional. It focuses on the last path planning involving rich contact between a multi-fingered hand and an object, which requires high precision and robustness against uncertainty. Given a pre-grasp and a goal functional grasp, the motion trajectory of this path starts from the pre-grasp and ends at the goal functional grasp. During this process, it is difficult to obtain the contact state of the hand-object from visual observation due to occlusion, so the joint torque located on each driving joint of the hand is taken as the force feedback information in this work.

The force feedback information, which dynamically updates due to contact collisions, can be used to sense the hand-object interaction state. If the multi-fingered hand automatically identifies and touches the functional part of the object while the object can still rest steadily with placement tolerances $(<0.01m)$, the motion trajectory of the functional grasp is considered successful. We further add disturbance $\epsilon$ $(<0.02m)$ to the initial horizontal position of the object to learn and verify its adaptability to uncertainty by utilizing the force feedback information.

We denote a pre-grasp as $G_0=(T_{0},R_{0},J_{0})$ and a goal grasp $G_g=(T_{g},R_{g},J_{g}) $, relative to the object at $({T^{obj}},{R^{obj}})$, where $J$ stands for the joint angle of the hand, and $T$ and $R$ stand for the position and orientation of the 6D pose. Our goal is to plan the motion from $G_0$ to $G_g$, guided by the force feedback of the hand joints. The movements of dexterous hands should be gentle to avoid moving objects.

\section{Method}

\subsection{Force-feedbacked Motion Planning via Reinforcement Learning}

The task of motion planning from pre-grasp to a goal functional grasp is challenging when relying solely on visual observations due to occlusion. Any errors in the hand or object positioning can result in knocking over or even destroying the object. To minimize such risks, we utilize force sensing in this work, which provides essential information about the local object geometry, contact forces, and grasp stability. Our motion planning strategy is based on joint torque feedback from each driving joint of the hand, which directly translates force cues into adaptive finger motion predictions. As the hand makes or breaks contact with the object, the set of joint torques applied to the object dynamically and quickly changes. We optimize the motion trajectory based on the observed joint torques, current hand pose, contact points on the object, and contact forces in order to predict the next desired hand pose and joint angles that will keep the object stable during grasping. As optimization progresses, the grasp becomes increasingly stable and plausible towards the goal functional grasp. Force feedback plays a crucial role in smoothly fitting the multi-fingered hand to the object surface to achieve a compliant grasp.

We train a deep reinforcement learning model using joint torques, hand pose, and joint angles of a multi-fingered hand as the observations and hand configuration changes as actions. An overview of the optimization process is presented in Figure \ref{fig_1}. To achieve a functional grasp of a given object at the end of the planning progress, we define a grasp reward function that reflects the grasp quality. Additionally, to minimize object movement when the hand touches it, we design another reward function that encourages the agent to move the object as little as possible. We employ the Soft Actor-Critic (SAC) algorithm \cite{haarnoja2018soft,laskin2020curl}, an off-policy reinforcement learning technique, to learn adaptive action policies through trial-and-error.

\subsection{Action and State Space}
\begin{table}[t]
	\caption{Observations Table}
	\label{table:table1} 
	\centering
	\begin{tabular}{c|c}
		\hline
		Observation & Dimension\\
		\hline
		Current joint torques of hand $F$   & 24 \\
		Current joint angles of hand $J$ & 24 \\
		Current 6D pose of hand  & 6 \\
		Joint motion controller $h$ & 1 \\
		Current driving joints of hand base  & 6 \\
		Previous 6D pose of hand  & 6 \\
		Previous joint motion controller  & 1 \\
		Goal 6D pose of hand $T_{g}, R_{g}$ & 6 \\
		Goal palm position & 3 \\
		\hline

	\end{tabular}
\end{table}

The action vector contains 24 joints controlling the fingers and wrist, as well as 6 joints controlling the hand base translation and rotation. All of these joints are controlled in position mode. The objective of the action control is to generate an optimized hand trajectory, which allows the hand to approach an object from a pre-grasp position to a goal functional grasp. We assume that the pre-grasp position is slightly more than 2 centimeters away from the surface of the object to avoid contact with the object due to initial positioning deviations, which can be obtained using a computer vision system and a grasp planner. The goal functional grasp is either obtained from recent static grasp synthesis methods \cite{zhu2021toward} or a grasp dataset \cite{wang2023dexgraspnet}. An example of the pre-grasp and goal grasp of a camera is shown in Figure \ref{fig_1}.

The task of motion planning from pre-grasp to a goal functional grasp is not a trivial one. Unlike the pick-and-place task that simply involves closing fingers, achieving a goal functional grasp, such as for dexterous tool use, requires precise positioning of the hand on the functional parts of an object. This task is further complicated by the occlusion of the object during hand-object interaction, making it difficult to determine the exact object position. In this work, we rely on the goal functional grasp and force feedback information to guide the hand motion. The reinforcement learning policy is designed such that the desired grasp pose drives the hand to achieve the desired pose on the object, while the joint torque feedback helps to locate the object, perceive its geometry, and provide compliant contact forces.

The entire multi-fingered hand is controlled by a 6-DoF hand base and one actuator per joint. The hand base globally adjusts the orientation and translation of the whole hand to make it reach the feasible approaching direction and distance to the object, while the actuator per joint locally controls the joint angle to force multiple fingers to contact the corresponding functional parts of the object. Since the hand leaves a small gap between the fingers and the object from a pre-grasp to a goal grasp, we employ a locally linear trajectory for each joint. This trajectory takes the pre-grasp joint angle $J_0$, goal joint angle $J_g$, and parameter $h$ as inputs and outputs the current joint angle as follows:

\begin{equation}
	J=h(J_{g}-J_{0})+J_{0},h \in [0,1]
\end{equation}

Here, $J_{g}-J_{0}$ denotes the range of joint angle, and $h \in [0,1]$ is an updated parameter that progressively controls the joint from the initial pre-grasp joint angle to the final goal joint angle. When $h=0$, the joint is at the pre-grasp angle, and when $h=1$, it is at the final goal angle. We use the same value of $h$ at each joint to synchronously drive multiple fingers towards the goal position.

The state vector consists of joint angles $J$, joint torques $F$ on each of the 24 driving joints, and the hand's 6D pose $(T, R)$, as detailed in TABLE \ref{table:table1}. To monitor the execution progress, the parameter $h$ that controls joint motion is also designed as an observation variable, which is defined as:

\begin{equation}
	h=1-0.5\max \limits_{i} \left| j_i -j_i^g \right|,\ j_i \in J,\ j_i^g \in J_{g}
\end{equation}

Here, $j_i$ and $j_i^g$ denote the current and goal joint angles for the $i$-th joint, respectively. In all joint angle changes, if the maximum absolute difference (normalized to [0,1]) between the current joint angles $J$ and the goal joint angles $J_g$ approaches 0, $h$ will be close to 1, indicating that all fingers have reached the goal hand pose. It is worth noting that we do not include any information about the object pose, geometry or mass in the state vector.

\begin{figure}[t]
	\centering
	\subfloat{\includegraphics[width=1.3in]{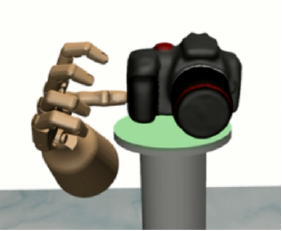}%
		\label{fig_2_1}}
	\hfil
	\subfloat{\includegraphics[width=1.3in]{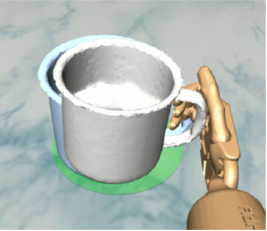}%
		\label{fig_2_2}}
	
	\caption{Terminal conditions: The hand moves in the incorrect orientation (Left). The object's displacement surpasses the predetermined limit (Right).}
	\label{fig_2}
\end{figure}
\begin{figure*}
	\centering
	\includegraphics[height=1.6in]{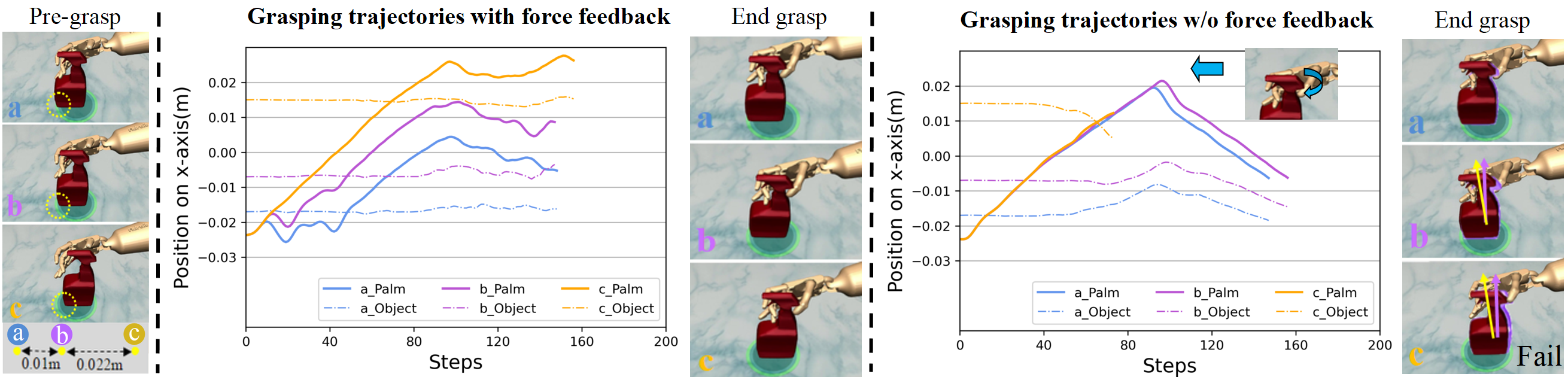}
	\caption{An example of the contribution of force feedback. Left: Pre-grasp. Middle: Grasping trajectories with force feedback. Right: Grasping trajectories without force feedback.}
	\label{fig_3}
\end{figure*}

\begin{figure}[t]
	\centering
	
	\subfloat{\includegraphics[width=2.5in]{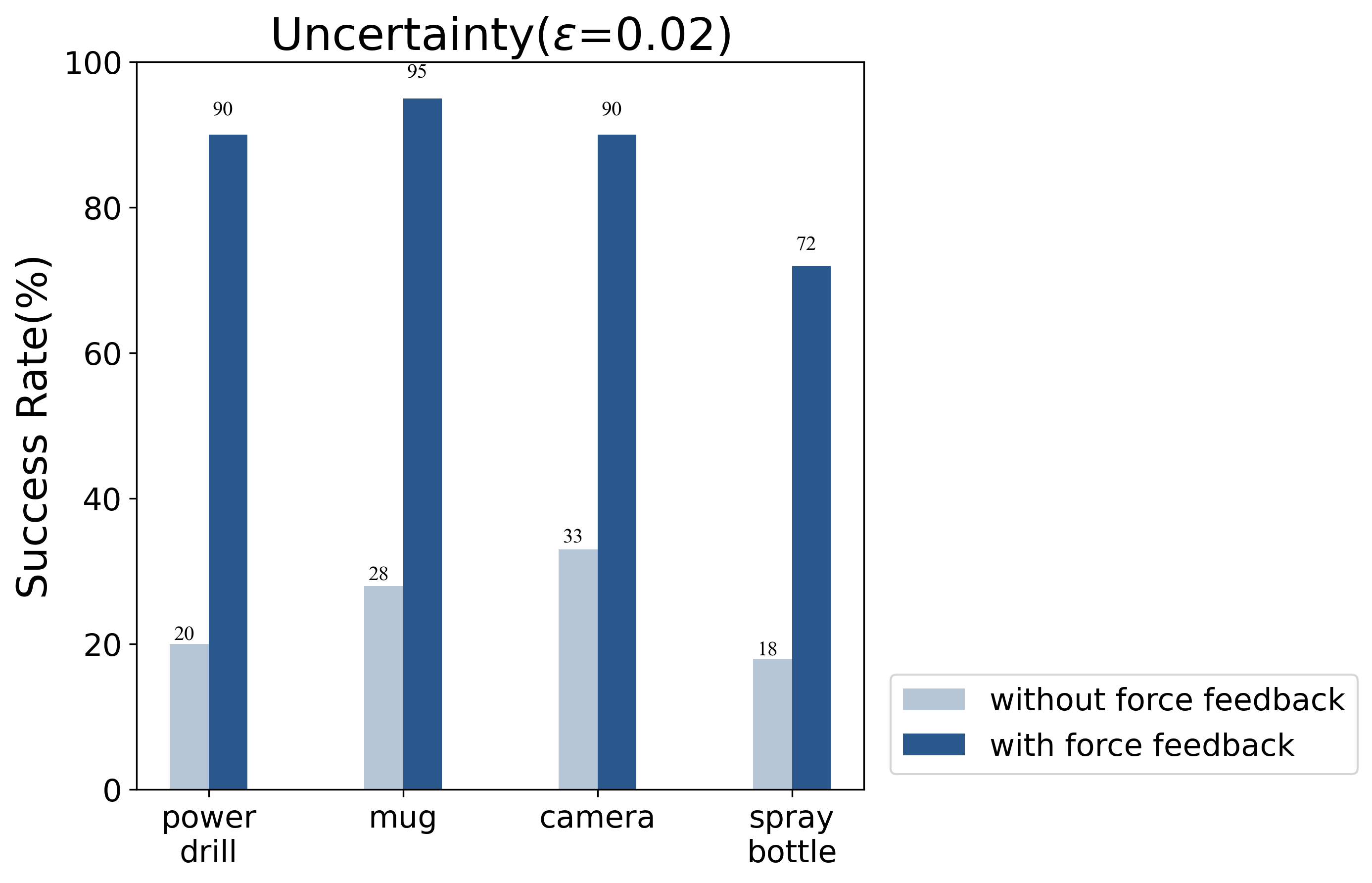}%
		\label{fig_second_case}}
	\caption{Quantitative analysis of the contribution of force feedback.}
	\label{fig_4}
\end{figure}

\subsection{Reward}
A carefully designed reward function is essential to guide the agent in learning the desired behavior. In our manipulation tasks, due to the initial uncertainty of the object's position, dexterous hands need to sense the real position of the object through touch. During this process, contact and collision between the hand and the object can cause further displacement of the object. Dexterous grasping requires reducing the movement of the object, allowing the hand to adjust to the object's position. Therefore, the ideal grasping result should be based on the relative position between the hand and the object in the goal grasp, achieving accurate functional grasping at the final position of the object. In order to achieve a functional grasp, we have designed a reward function $r_a$ that guides the robot's joint angles towards the goal pose. Specifically, the observation variable $h$ indicates how close the joint angles are to the goal angles, and the closer they are, the greater the reward obtained.

\begin{equation}
	\label{reward_b}
	r_a=-\left|\ 1- h \ \right|
\end{equation}

We further design the reward function $r_b$ to guide the palm to the goal grasping position. Due to the multi-point contact constraints between the hand and the object, we only use the position of a single point on the palm to guide its movement. We obtain the current position of the palm relative to the object from the simulation system, denoted as $\textbf{p}$. The reward function is used to guide the palm to minimize the deviation between its current position $\textbf{p}$ and the goal position $\textbf{p}_{g}$. It should be noted that the variable $\textbf{p}_{g}$ here denotes the desired position of the palm relative to the current object position, as opposed to the goal palm position provided in the state, which is defined relative to ${T^{obj}}$. This approach has shown promising results in simulation experiments.

\begin{equation}
	r_b=-\Vert	\textbf{p}-{\textbf{p}_{g}} \Vert_2
\end{equation}

Moreover, we employ $r_c$ to incentivize the agent to minimize object displacement during operation, which is calculated as the negative distance between the object positions before and after the operation:

\begin{equation}
	r_c=-\Vert {{T^{obj}_{c}}}-{\widetilde T^{obj}}\Vert_2
\end{equation}
\begin{equation}
	{\widetilde T^{obj}}={T^{obj}}+\epsilon
\end{equation}

Here, ${T^{obj}_{c}}$ denotes the current object position, and ${\widetilde T^{obj}}$ represents the initial position of the object with added uncertainty.

Finally, we use the weight coefficients to scale the reward functions mentioned above, approximating a range of [-1,0]. The total reward is defined as:

\begin{equation}
	r=\omega+\omega_a r_a+\omega_b r_b+\omega_c r_c
\end{equation}
where $\omega=1.5$, $\omega_a=1$, $\omega_b=20$, and $\omega_c=100$.

\subsection{Terminal Conditions}
It is crucial to encourage the agent to focus on important regions of the state space and avoid unnecessary exploration. We define successful task completion based on three conditions: when the object's movement is less than $0.01m$ ($\left|r_c\right|<0.01$), the final joint angles of the hand are 95$\%$ close to the goal angles ($\left|r_a\right|<0.05$), and the deviation of the palm position is less than $0.01m$ ($\left|r_b\right|<0.01$). Once these conditions are met, the agent receives an additional reward of $1000$. Moreover, a reward of $1.5*(200-step)$ is designed to incentivize the agent to complete the task in as few steps as possible, where the maximum number of steps allowed to execute a task is 200. The values of $r_a$ and $r_c$ are continuously monitored in real-time. If the object's movement exceeds $0.01m$ or the palm position deviation surpasses $0.1m$, the grasp fails immediately, as shown in Figure \ref{fig_2}. In addition, if the total reward falls below $-50$, indicating inefficient exploration, the task will be aborted promptly.
\begin{figure*}
	\centering
	\includegraphics[width=5.4in]{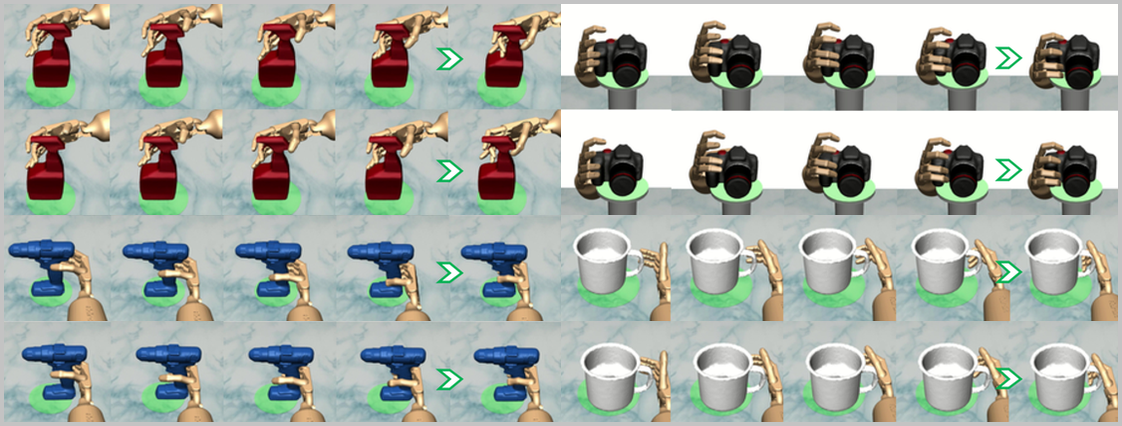}
	\caption{Guided by real-time force feedback, diverse paths are continually adjusted, aiming to achieve precise grasps while minimizing object displacement.}
	\label{fig_5}
\end{figure*}

\begin{figure*}
	\centering
	\includegraphics[width=6.1in]{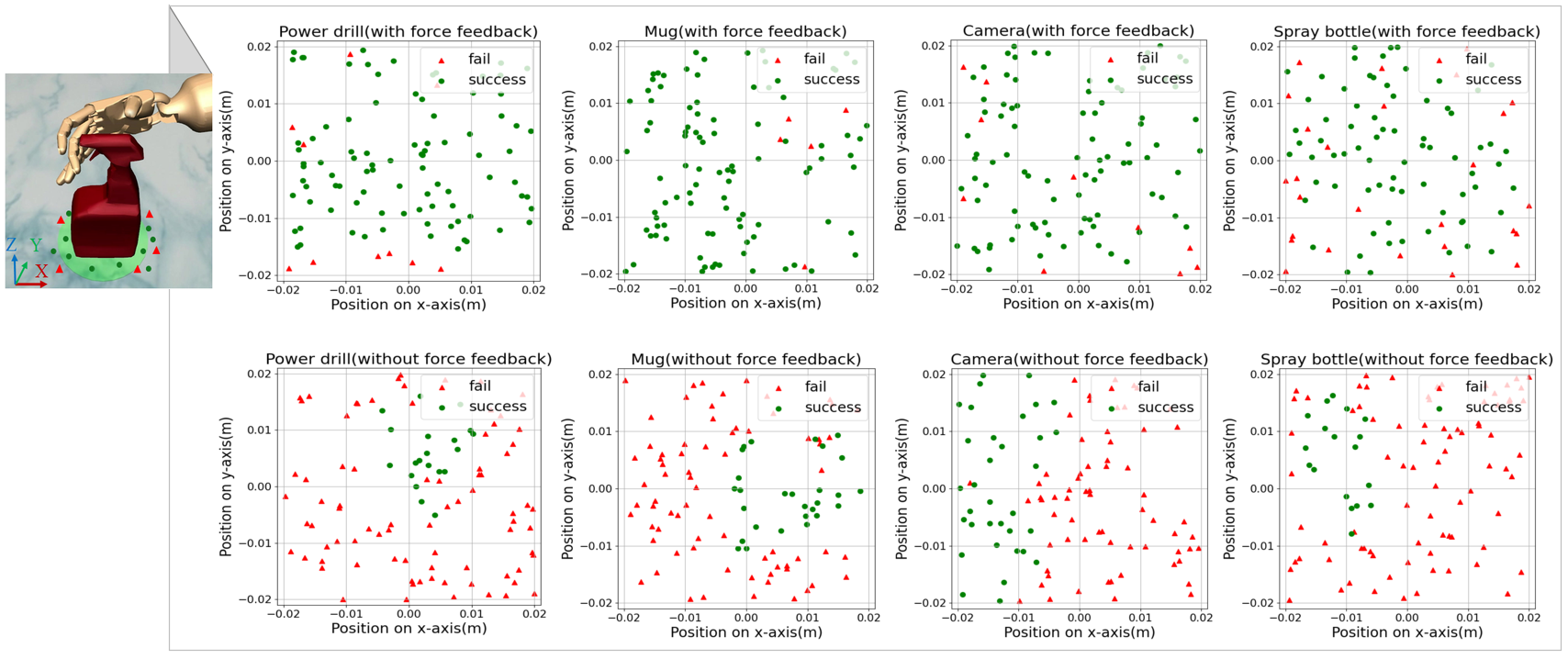}
	\caption{Adaptation to object location.}
	\label{fig_6}
\end{figure*}

\section{Experiments}
\subsection{Simulation Experiment Settings}
We perform experiments using the MuJoCo physics simulator \cite{todorov2012mujoco} and the ADROIT dexterous hand \cite{kumar2013fast}, which is a 24-DoF anthropomorphic platform equipped with six driving joints on the hand base, allowing the hand to move freely in space. As our study focuses on the adaptive actions guided by force feedback, factors such as tendon elasticity and weak driving forces could hinder the perception of force feedback. Therefore, we make modifications to the platform, including removing hand tendons, increasing the range of joint forces, and building a position controller to enhance controllability.

Four tasks are designed to achieve functional grasping, demanding precise pose matching between the hand and the object. The pre-grasp hand pose and the goal grasp hand pose of the four objects are labeled according to \cite{zhu2021toward}.

During training, the simulation system is reset when the termination condition is reached, and the total number of learning steps is 1000000. The first 1000 steps adopt a random policy, and then a batch of data is randomly selected from the buffer to update the policy in each step. The impact of initial random exploration on the training process and outcomes is overcome through repeated experiments. We test the policies learned by the agents every 2000 steps and retain the five best policies during the entire learning process. Finally, the best action policy is selected through 100 random tests. This learning process takes approximately 4 hours on an RTX2080 Ti. 

\subsection{Experimental Results}

\textbf{Contribution of force feedback}. Without visual sensing, a five-fingered robot hand that relies solely on force feedback to learn robust motion and precise grasp is quite challenging, as objects are prone to displacement by touching with force, and the lighter the object, the greater the displacement. To avoid significant displacement of an object, we first increase its weight by 5 times to facilitate the policy training of motion planning, then verify the effectiveness of the learned policy at its original or lighter weight.

To reveal the role of force feedback, we take grasping a spray bottle as an example,
as shown in Figure \ref{fig_3}. Three grasps initiate from their pre-grasps with varying initial object positions: (-0.017, 0.01), (-0.007, 0.01), and (0.015, 0.01), respectively, in scenarios a, b, and c (depicted on the left side of Figure \ref{fig_3}). The grasping trajectories of the X-coordinate of the palm (solid
line) and the object (dashed line) are depicted, followed by end grasp images. It
can be observed that when force feedback is present, the hand adjusts its
trajectories under the guidance of force feedback to generate different grasping
actions with small object movement (in the middle of Fig. \ref{fig_3}). On the contrary, in the
absence of force feedback, the dexterous hand performs similar grasping trajectories
for scenes a and b, resulting in a large degree of object movement, and the grasping
motion fails for scene c (on the right of Fig. \ref{fig_3}). This demonstrates that the real-time
force feedback during hand-object interaction is crucial for grasp motion planning.
Figure \ref{fig_4} presents the quantitative results that achieve a grasp success rate of $90\%$
for most objects with force feedback, while the success rate is below $35\%$ without force
feedback.

Figure \ref{fig_5} further illustrates the motion planning of the dexterous hand as it
performs functional grasping of four distinct objects guided by force feedback. The
figure highlights the adaptability of our method in generating diverse grasping
trajectories and grasp types tailored to different initial object positions(as indicated by the green circle). These diverse
paths are continually adjusted in real-time as the hand interacts and collides with the
objects, aiming to achieve precise grasps while minimizing object displacement. It
can be observed that the generated grasps are very similar to those of humans, exhibiting dexterity, flexibility, and accuracy. We also submit a video as supplementary
materials to demonstrate grasping with force feedback.

\textbf{Adaptation to object location}. 
To verify the ability of force feedback
to help grasp planning adapt to the changes in the object’s location in more
experiments, one hundred random tests are conducted on each of the objects using the
learned grasping policy, and the results are shown in Figure \ref{fig_6}. For each test, the
initial X and Y coordinates of the object are randomly selected within a range of $\pm2$ $cm$
 and plotted on the chart. A successful grasp is marked with a green circle, while a
failed grasp is marked with a red triangle. It can be observed from the distribution map that
most of the grasps are successful in the presence of force feedback.
Conversely, successful grasps occur at relatively fewer locations in the absence of
force feedback.

\begin{figure}[t]
	\centering
	\subfloat{\includegraphics[width=2.5in]{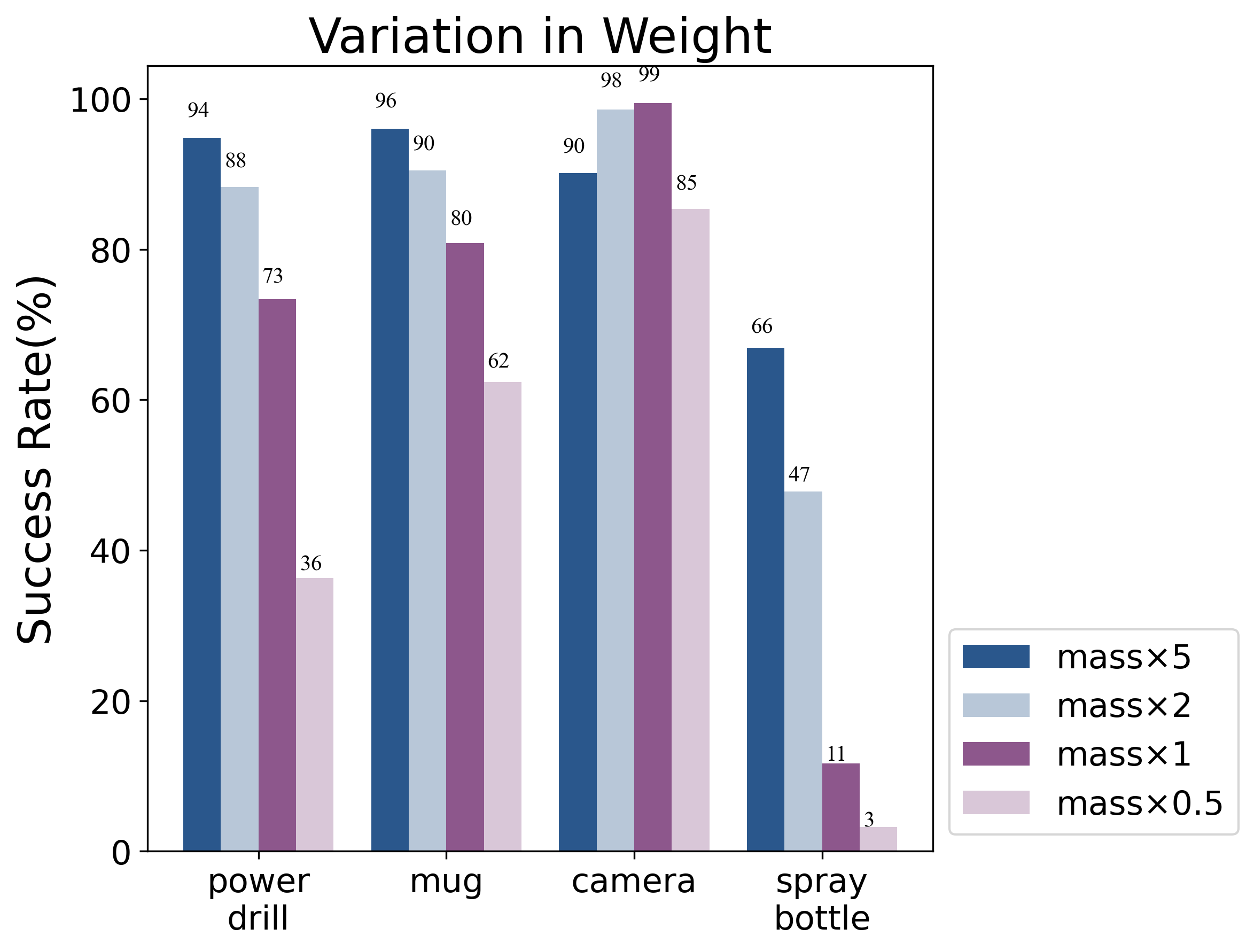}%
              }
	\caption{Grasping lighter objects.}
	\label{fig_7}
\end{figure}

\textbf{Grasping lighter objects}. 
Objects of the same
shape and function, regardless of weight, have the same grasp. The grasping policies
learned from the previously weighted objects are still applicable to non-weighted or
light objects. The only difference is that the latter objects are more easily knocked
away due to their lighter weight. Accordingly, we expand the permissible range of object movement during manipulation from 0.01m to 0.03m and reiterate the grasping procedure on the objects, utilizing the previously acquired policy. Figure \ref{fig_7} shows the average success rate over
10 trials. It is evident that high success rates were achieved for grasping the power drill, mug, and camera at their original weights, with respective rates of $73\%$, $80\%$, and $99\%$, excluding the spray bottle. For objects with half the weight, the success rate
decreases accordingly, yet still exceeds $50\%$ on mug and camera. This indicates that
the grasping policy of a dexterous hand can be learned through the force sensing. We
speculate that the lower success rates of power drill and spray bottle may be due to
their relatively complex grasping poses, which make them more prone to collisions,
resulting in failed grasps.

\begin{figure}[t]
	\centering
	
	\subfloat{\includegraphics[width=2.4in]{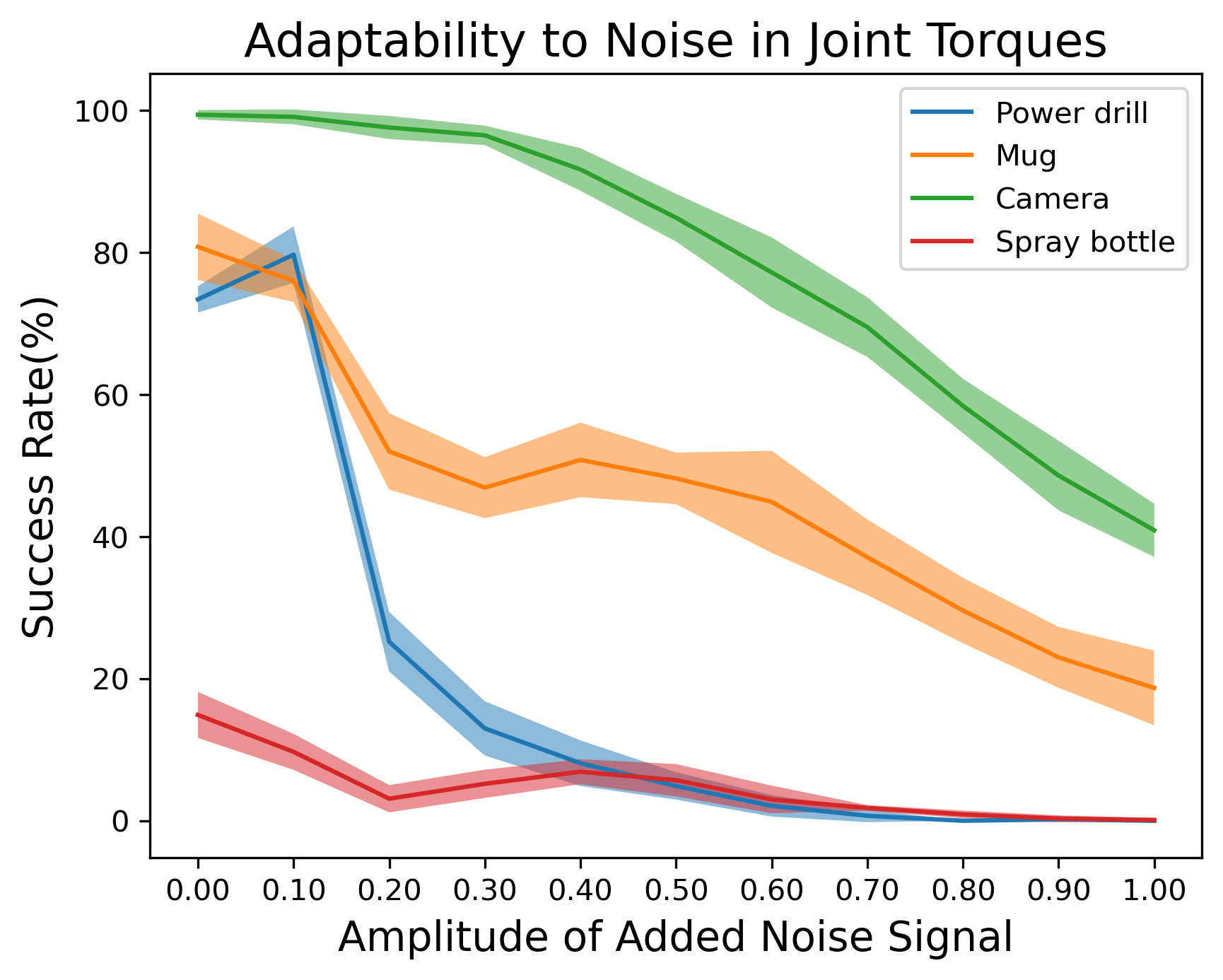}%
		}
	\caption{Robustness against joint torque noise.}
	\label{fig_8}
\end{figure}

\textbf{Robustness against joint torque noise}.
To evaluate the performance of the
grasping policy against joint torque noise, we conduct simulated experiments
introducing joint torque noise ranging from 0 to 1 Nm in increments of 0.1 Nm.
Figure \ref{fig_8} illustrates the success rates across 10 trials for each noise level. Notably, the
camera policy maintains an impressive success rate of $80\%$ even with noise
interference up to 0.5 Nm, and the policies of the mug and power drill also achieve
a high success rate within a tolerance of 0.1 Nm. The experiments indicate the
robustness of our grasping policy.

\subsection{Comparison with Prior Methods}

We have summarized representative papers considering joint torque feedback on
five-fingered dexterous hands, as shown in TABLE \ref{table:table2}. Both \cite{chen2022towards} and \cite{xu2023unidexgrasp} assume
accurate visual tracking of the object, and the role of joint torque is not validated. \cite{liang2021multifingered} is the closest to our work, making hand-object
interactions rely on joint torque feedback. It demonstrates
that tactile perception-based joint torque feedback enhances
grasp performance. However, this method does not adjust the wrist position after the hand reaches the pre-grasp position. In contrast, our proposed method offers
unrestricted arm movement that is crucial for achieving dexterous grasp.

\begin{table}[t]
	\caption{Comparison with Prior Methods}
	\label{table:table2} 
	\centering
	\begin{tabular}{c|c|c|c|c}
		\hline
		 & \cite{chen2022towards}&\cite{xu2023unidexgrasp}&\cite{liang2021multifingered}& Ours\\
		\hline
		 Vision &   \checkmark  & \checkmark   & $\times$  & $\times$   \\
		Precise Grasping &  $\times$ & $\times$ & $\times$ & \checkmark \\
		Role of Joint Torque  &  - & -	 & \checkmark & \checkmark \\
		
		\hline

	\end{tabular}
\end{table}

\section{Conclusion And Future Work}

We have introduced a novel adaptive motion planning method for multi-fingered functional grasping utilizing force feedback and analyzed its performance on four distinct objects. The reinforcement learning model established is capable of learning policies to adjust actions in real-time based on joint torque feedback, enabling adaptation to disturbances in initial object poses. The achieved dexterous grasps demonstrate the capabilities of robot manipulation that emulate human dexterity. However, our method still has limitations. For instance, when dealing with complex grasps, such as those involving spray bottles, relying solely on joint torque feedback may not suffice to learn adaptive grasping. Here, the addition of real-time visual feedback becomes crucial. In our future work, we intend to integrate joint torque perception with multimodal sensing, such as vision and touch, aiming to enhance the robot's adaptability and dexterity across a diverse range of grasping tasks.

\bibliographystyle{IEEEtran}
\bibliography{reference}

\end{document}